\documentclass{article}

\usepackage{microtype}
\usepackage{graphicx}
\usepackage{subcaption}
\usepackage{booktabs} % for professional tables
\usepackage{enumitem}

\usepackage{hyperref}
\usepackage{svg}

\usepackage[preprint]{icml2026}

\usepackage{amsmath}
\usepackage{amssymb}
\usepackage{mathtools}
\usepackage{amsthm}
\usepackage{bm} % For bold math symbols
\usepackage[capitalize,noabbrev]{cleveref}

\theoremstyle{plain}

\theoremstyle{definition}

\theoremstyle{remark}

\usepackage[textsize=tiny]{todonotes}

\icmltitlerunning{Improving Discrete Optimisation Via Decoupled Straight-Through Estimator}

\begin{document}

\twocolumn[
  % temporary for now:
  \icmltitle{Improving Discrete Optimisation Via Decoupled Straight-Through Estimator}

  % It is OKAY to include author information, even for blind submissions: the
  % style file will automatically remove it for you unless you've provided
  % the [accepted] option to the icml2026 package.

  % List of affiliations: The first argument should be a (short) identifier you
  % will use later to specify author affiliations Academic affiliations
  % should list Department, University, City, Region, Country Industry
  % affiliations should list Company, City, Region, Country

  % You can specify symbols, otherwise they are numbered in order. Ideally, you
  % should not use this facility. Affiliations will be numbered in order of
  % appearance and this is the preferred way.
  \icmlsetsymbol{equal}{*}

  \begin{icmlauthorlist}
    \icmlauthor{Rushi Shah}{nus}
    \icmlauthor{Mingyuan Yan}{nyu}
    \icmlauthor{Michael Curtis Mozer}{gdm}
    \icmlauthor{Dianbo Liu}{nus}
  \end{icmlauthorlist}

  \icmlaffiliation{nus}{College of Design and Engineering, National University of Singapore, Singapore}
  \icmlaffiliation{gdm}{Google Deepmind, California, USA}
  \icmlaffiliation{nyu}{New York University, USA}

  \icmlcorrespondingauthor{Dianbo Liu}{dianbo@nus.edu.sg}

  % You may provide any keywords that you find helpful for describing your
  % paper; these are used to populate the "keywords" metadata in the PDF but
  % will not be shown in the document
  \icmlkeywords{Straight-Through Estimator, Discrete Optimisation, Gradient Estimation, Temperature Scaling, Categorical Variables}

  \vskip 0.3in
]

% this must go after the closing bracket ] following \twocolumn[ ...

% This command actually creates the footnote in the first column listing the
% affiliations and the copyright notice. The command takes one argument, which
% is text to display at the start of the footnote. The \icmlEqualContribution
% command is standard text for equal contribution. Remove it (just {}) if you
% do not need this facility.

% Use ONE of the following lines. DO NOT remove the command.
% If you have no special notice, KEEP empty braces:
\printAffiliationsAndNotice{}  % no special notice (required even if empty)
% Or, if applicable, use the standard equal contribution text:
% \printAffiliationsAndNotice{\icmlEqualContribution}

% to be revised after other sections are complete
\begin{abstract}
The Straight-Through Estimator (STE) is the dominant method for training neural networks with discrete variables, enabling gradient-based optimisation by routing gradients through a differentiable surrogate. However, existing STE variants conflate two fundamentally distinct concerns: forward-pass stochasticity, which controls exploration and latent space utilisation, and backward-pass gradient dispersion i.e how learning signals are distributed across categories. We show that these concerns are qualitatively different and that tying them to a single temperature parameter leaves significant performance gains untapped. We propose Decoupled Straight-Through (Decoupled~ST), a minimal modification that introduces separate temperatures for the forward pass ($\tau_f$) and the backward pass ($\tau_b$). This simple change enables independent tuning of exploration and gradient dispersion. Across three diverse tasks (Stochastic Binary Networks, Categorical Autoencoders, and Differentiable Logic Gate Networks), Decoupled~ST consistently outperforms Identity~STE, Softmax~STE, and Straight-Through Gumbel-Softmax. Crucially, optimal $(\tau_f, \tau_b)$ configurations lie far off the diagonal $\tau_f = \tau_b$, confirming that the two concerns do require different answers and that single-temperature methods are fundamentally constrained.
\end{abstract}

\begin{figure*}[ht]
    \centering
    \includegraphics[width=\textwidth]{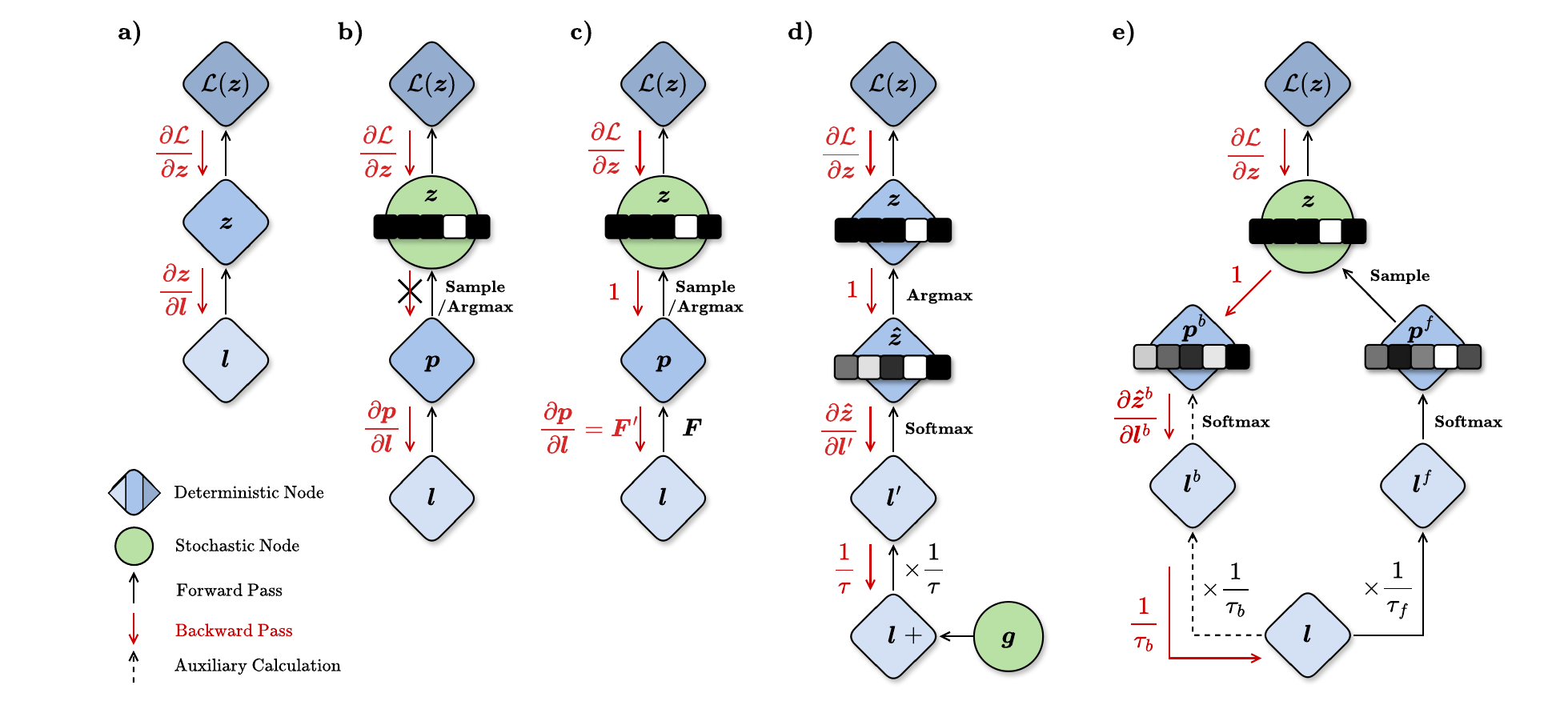}
    \caption{\textbf{Gradient Flow Comparison for different gradient estimation methods.} \textbf{(a)} In the continuous setting, node \( \bm{z} \) is a deterministic variable, and gradients can be propagated back through \( \bm{z} \) and \( \mathcal{L}(\bm{z}) \) directly using the chain rule. \textbf{(b)} When \( \bm{z} \) represents a discrete categorical variable, the sampling/argmax process from \( \bm{p} \) breaks the backpropagation path. \textbf{(c)} Standard STE, where $\partial \bm{z} / \partial \bm{p}$ is approximated as 1 during the backward pass, allowing gradients to flow through non-differentiable stochastic nodes. Different choices for $\boldsymbol{F}$ yields different STE variants. \textbf{(d)} ST-GS with temperature, scaling logits by a single temperature \( \tau \) for both forward and backward passes after injecting stochasticity by adding a Gumbel noise sample. \textbf{(e)} Decoupled ST uses separate temperatures: $\tau_f$ controls forward-pass stochasticity while $\tau_b$ controls backward-pass gradient dispersion.}
    \label{fig:main-figure}
\end{figure*}

\section{Introduction}
Discrete representations serve as a powerful inductive bias in deep learning, enhancing systematic generalisation \citep{liu2021discrete}, improving compression efficiency \citep{balle2016end}, and promoting interpretability through natural factor disentanglement \citep{chen2016infogan}. They appear throughout modern architectures: in discrete latent variable models such as VQ-VAEs \citep{van2017neural} and categorical autoencoders \citep{jang2016categorical}; in world models for sample-efficient planning \citep{hafner2020mastering, micheli2022transformers}; in high-fidelity image generation \citep{chang2022maskgit, tian2024visual}; and in efficient, interpretable networks such as Stochastic Binary Networks \citep{courbariaux2015binaryconnect} and Differentiable Logic Gate Networks \citep{petersen2022deep}.

Training models with discrete variables, however, poses a fundamental challenge: non-differentiability. Several strategies have emerged to enable gradient-based optimisation: policy-gradient methods such as REINFORCE \citep{Williams2004SimpleSG}, relaxation-based approaches like the Gumbel-Softmax trick \citep{jang2016categorical, maddison2016concrete}, and the Straight-Through Estimator (STE) \citep{bengio2013estimating, hinton2012lecture15d}. The STE routes gradients through a differentiable surrogate function during backpropagation \citep{shekhovtsov2021reintroducing}; different choices of this surrogate yield widely-used variants such as the Identity STE, Softmax STE, and Straight-Through Gumbel-Softmax (ST-GS). Due to its simplicity and effectiveness, the STE family has become the dominant approach for training discrete models.

Existing STE methods, however, overlook a core consideration: optimising discrete variables requires answering \emph{two distinct questions}. First, \emph{what level of stochasticity should govern the forward pass?} Higher stochasticity encourages exploration, improves latent space utilisation, and prevents the model from ``cheating by encoding information in floating-point operations''; but too much destabilises training. Second, \emph{how should gradients be distributed across categories in the backward pass?} Concentrating gradients on the selected category yields strong learning signals but risks ``dead'' categories, while spreading gradients too broadly dilutes the signal. Current approaches either conflate both questions or address only one, leaving substantial performance gains on the table.

To disentangle these concerns, we propose \emph{Decoupled Straight-Through} (Decoupled ST), a minimal modification that introduces separate temperatures for the forward pass ($\tau_f$) and the backward pass ($\tau_b$). We evaluate Decoupled ST on Stochastic Binary Networks, Categorical Autoencoders, and Differentiable Logic Gate Networks, demonstrating consistent improvements over Identity STE, Softmax STE, and ST-GS. Notably, the optimal $(\tau_f, \tau_b)$ configurations lie far off the diagonal $\tau_f = \tau_b$, confirming that the two questions indeed require different answers. We further analyse the distinct effects of each temperature, showing that $\tau_f$ mediates a trade-off between exploration and gradient variance while $\tau_b$ controls gradient dispersion and magnitude.

% [noitemsep, topsep=0pt]
Our contributions are as follows:
\begin{itemize}
\item We identify a fundamental limitation of existing STE methods: they conflate forward-pass stochasticity (which controls exploration) with backward-pass gradient shaping (which controls learning signals).
\item We propose \emph{Decoupled ST}, a minimal modification that introduces separate forward and backward temperatures, and demonstrate consistent improvements across a diverse set of experiments.
\item We provide empirical analysis showing that both temperatures control qualitatively different aspects of training dynamics, and that optimal configurations lie off the diagonal.
\end{itemize}

\section{Preliminaries and Related Work}

\subsection{Discrete Variables and Gradient Computation}

Consider a computational graph in which a discrete choice must be made over $k$ categories. Let $\bm{l} = (l_1, l_2, \dots, l_k) \in \mathbb{R}^k$ denote \emph{unnormalized logits} produced by upstream parameters $\theta$. To select a category, we first compute probabilities via a function $F$ (typically softmax, $\bm{p} = F(\bm{l}) = \text{softmax}(\bm{l})$) and then obtain a one-hot encoding:
\begin{equation}
    \bm{z} \in \{0, 1\}^k, \quad \sum_{i=1}^{k} z_i = 1,
\end{equation}
either by taking the argmax ($\bm{z} = \text{one-hot}(\arg\max_i\, p_i)$) or by sampling ($\bm{z} \sim \text{Categorical}(\bm{p})$). The discrete $\bm{z}$ then flows downstream, ultimately contributing to a scalar loss $\mathcal{L}$.

Optimising $\theta$ by gradient descent requires computing $\partial \mathcal{L} / \partial \bm{l}$. However, when the forward pass includes discrete selection, whether through argmax or sampling, the chain rule fails: the Jacobian $\partial \bm{z} / \partial \bm{l}$ is either zero almost everywhere (for argmax) or undefined (for sampling). This breaks gradient propagation entirely (Figure~\ref{fig:main-figure}b).

One might hope to compute the exact expected gradient by brute force, but this requires summing over all $k$ possible discrete states, each demanding a separate forward and backward pass, making this intractable for all but the smallest $k$. For models with multiple categorical choices (e.g., $n$ independent $k$-way variables), the cost grows exponentially as $k^n$. Alternatively, REINFORCE-style policy-gradient estimators \citep{Williams2004SimpleSG} provide unbiased gradient estimates:
\begin{equation}
    \nabla_{\bm{l}} \mathbb{E}_{\bm{z} \sim p(\cdot | \bm{l})}[\mathcal{L}(\bm{z})] = \mathbb{E}_{\bm{z}}\left[ \mathcal{L}(\bm{z}) \nabla_{\bm{l}} \log p(\bm{z} | \bm{l}) \right].
\end{equation}
However, single-sample REINFORCE estimates suffer from high variance; reducing this variance requires many samples, each incurring the cost of a full forward and backward pass, a computational burden that makes REINFORCE difficult to apply in large-scale settings.

\subsection{Straight-Through Estimators}

As an alternative to Monte Carlo estimation, the Straight-Through Estimator (STE) \citep{bengio2013estimating} bypasses the non-differentiable sampling step entirely by constructing a \emph{surrogate gradient path} \citep{shekhovtsov2021reintroducing}. The core idea is simple: in the forward pass, we make a discrete decision as usual; in the backward pass, we pretend the discretization never happened and route gradients through a differentiable surrogate function $F: \mathbb{R}^k \to \mathbb{R}^k$ that transforms the logits (Figure~\ref{fig:main-figure}c).

Algorithm~\ref{alg:ste} formalizes this procedure. Given logits $\bm{l}$, the forward pass applies the surrogate $F$ to obtain continuous values $\bm{p} = F(\bm{l})$, then discretizes via argmax or sampling to produce $\bm{z}$. The backward pass ignores this discretization and computes gradients as if $\bm{z} = \bm{p}$:
\begin{equation}
    \frac{\partial \mathcal{L}}{\partial \bm{l}} \approx \frac{\partial \mathcal{L}}{\partial \bm{z}} \frac{\partial F(\bm{l})}{\partial \bm{l}}.
\end{equation}
Different choices of the surrogate $F$ yield different STE variants, each with its own ways to reweigh the gradient belonging to different activation regions.

\begin{algorithm}[t]
\caption{Straight-Through Estimator}
\label{alg:ste}
\begin{algorithmic}[1]
\STATE \textbf{Input:} logits $\bm{l} \in \mathbb{R}^k$, surrogate function $F$
\STATE \textbf{Forward:}
\STATE \quad $\bm{p} \leftarrow F(\bm{l})$ \hfill $\triangleright$ \emph{apply surrogate}
\STATE \quad $\bm{z} \leftarrow \text{discretize}(\bm{p})$ \hfill $\triangleright$ \emph{argmax or sample}
\STATE \quad \textbf{return} $\bm{z}$
\STATE \textbf{Backward:} given $\frac{\partial \mathcal{L}}{\partial \bm{z}}$
\STATE \quad \textbf{return} $\frac{\partial \mathcal{L}}{\partial \bm{l}} \leftarrow \frac{\partial F(\bm{l})}{\partial \bm{l}}^\top \frac{\partial \mathcal{L}}{\partial \bm{z}}$ \hfill $\triangleright$ \emph{straight-through}
\end{algorithmic}
\end{algorithm}

\subsection{Common STE Variants}

The general STE framework (Algorithm~\ref{alg:ste}) encompasses several widely-used methods, each corresponding to a particular choice of surrogate $F$.

\paragraph{Identity STE \citep{bengio2013estimating}.} The simplest variant sets $F(\bm{l}) = \bm{l}$ (the identity function) and uses argmax for discretization: $\bm{z} = \text{one-hot}(\arg\max_i\, l_i)$. Since $\partial F / \partial \bm{l} = \bm{I}$, gradients pass through unchanged: $\partial \mathcal{L}/\partial \bm{l} = \partial \mathcal{L}/\partial \bm{z}$.

\paragraph{Softmax STE \citep{hinton2012lecture15d}.} This variant sets $F(\bm{l}) = \text{softmax}(\bm{l}/\tau)$, where $\tau > 0$ is a temperature parameter. The discretization step samples from the resulting categorical distribution: $\bm{z} \sim \text{Categorical}(\bm{p})$. Both the forward-pass sampling distribution and the backward-pass surrogate Jacobian are governed by the same temperature $\tau$.

\paragraph{Straight-Through Gumbel-Softmax (ST-GS) \citep{jang2016categorical, maddison2016concrete}.} Building on the Gumbel-Max reparameterization, ST-GS first perturbs the logits with i.i.d.\ Gumbel noise: $\tilde{\bm{l}} = \bm{l} + \bm{g}$, where $g_i \sim \text{Gumbel}(0,1)$. The surrogate is $F(\tilde{\bm{l}}) = \text{softmax}(\tilde{\bm{l}}/\tau)$, and discretization uses argmax: $\bm{z} = \text{one-hot}(\arg\max_i\, \tilde{l}_i)$. The argmax over Gumbel-perturbed logits is equivalent to sampling from $\text{Categorical}(\text{softmax}(\bm{l}))$, so the forward pass produces categorical samples regardless of $\tau$ (Figure~\ref{fig:main-figure}d).

\section{The Conflation Problem}
\label{sec:conflation}

Training models with discrete representations raises two fundamental questions that influence training dynamics in distinct ways.

\paragraph{Q1: How much stochasticity is needed in the forward pass?} High stochasticity directly increases the exploration capabilities of the model, leading to higher latent space utilisation. However, excessive stochasticity destabilises training because it indirectly increases gradient variance, worsening the gradient signal-to-noise ratio. Conversely, low stochasticity promotes stable training by reducing gradient variance, but risks poor utilisation of the latent space, eventually bottlenecking the expressivity of the model. Several factors influence where the optimal stochasticity level lies: how robustly the network before the categorical bottleneck can handle variance in gradients; how much stochasticity the network after the bottleneck can tolerate; and how much latent space utilisation is needed for a given task: when model capacity far exceeds dataset complexity, gradient stability may matter more than utilisation, but when capacity is tight, fully utilising the discrete bottleneck may outweigh concerns about gradient variance.

\paragraph{Q2: How should gradients be distributed across categories in the backward pass?} Concentrating gradients on the highest-probability category provides strong, informative learning signals but may cause ``dead categories'' that never receive updates and hence never get activated. This makes the network inflexible: once a category is selected, the model receives little information about whether other options might have been better. On the other hand, distributing gradients too evenly provides some signal to alternative categories, but that signal becomes too weak and diluted for meaningful learning, leading to very slow training.

\subsection{How Current Methods Handle These Questions}

Existing STE methods either conflate these questions into a single control or ignore them entirely.

\paragraph{Softmax STE} uses a single temperature $\tau$ that governs both the forward-pass sampling entropy and the backward-pass gradient fidelity. Lowering $\tau$ reduces forward stochasticity \emph{and} concentrates gradients on selected categories; raising $\tau$ increases stochasticity \emph{and} disperses gradients. This coupling forces a compromise between exploration and gradient quality, inevitably leaving performance on the table.

\paragraph{Identity STE} ignores both questions entirely. It uses deterministic argmax selection and passes gradients unchanged through the identity Jacobian. While straightforward and temperature-free, the Identity STE ignores the probabilistic structure entirely, providing a crude approximation that can lead to poor gradient quality.

\paragraph{ST-GS} partially decouples the concerns but in a limited way. The Gumbel noise injection makes forward stochasticity uncontrollable: the forward pass always produces categorical samples with full sampling variance. The temperature $\tau$ controls only the backward-pass gradient fidelity. Furthermore, ST-GS computes gradients with respect to noise-perturbed logits $\tilde{\bm{l}} = \bm{l} + \bm{g}$ rather than the true logits $\bm{l}$, introducing an additional source of approximation error.

This analysis reveals that no existing method provides independent control over both forward-pass stochasticity and backward-pass gradient distribution.

\section{Methodology}
\label{sec:methodology}

We propose \emph{Decoupled Straight-Through} (Decoupled ST), a simple modification that introduces separate temperatures for the forward and backward passes, enabling independent control over both concerns (Figure~\ref{fig:main-figure}e).

\subsection{Formulation}

Let $\bm{l} \in \mathbb{R}^k$ denote logits and define the temperature-scaled softmax:
\begin{equation}
p_i(\tau) = \frac{\exp(l_i / \tau)}{\sum_{j=1}^{k} \exp(l_j / \tau)}.
\end{equation}
In Decoupled ST, the forward pass samples from $\bm{p}^f = \text{softmax}(\bm{l} / \tau_f)$:
\begin{equation}
    \bm{z} \sim \text{Categorical}(\bm{p}^f).
\end{equation}
Since $\bm{z}$ is discrete, we cannot directly compute $\partial \mathcal{L} / \partial \bm{l}$. The STE approximates the expected gradient by routing through a differentiable surrogate. In Decoupled ST, we use a separate backward temperature $\tau_b$ to define $\bm{p}^b = \text{softmax}(\bm{l} / \tau_b)$, and the gradient estimate becomes:
\begin{equation}
    \hat{g}(\bm{l}) = J(\tau_b)^\top \frac{\partial \mathcal{L}(\bm{z})}{\partial \bm{z}}, \quad \text{where } \bm{z} \sim \text{Categorical}(\bm{p}^f),
    \label{eq:decoupled-gradient}
\end{equation}
where $J(\tau_b) = \frac{1}{\tau_b}\big( \text{diag}(\bm{p}^b) - \bm{p}^b {\bm{p}^b}^\top \big)$ is the softmax Jacobian at temperature $\tau_b$. The key insight is that the sample $\bm{z}$ is drawn from $\bm{p}^f$ (controlled by $\tau_f$), while the gradient is routed through $J(\tau_b)$ (controlled by $\tau_b$). Algorithm~\ref{alg:decoupled-st} formalizes this procedure.

\begin{algorithm}[t]
\caption{Decoupled Straight-Through Estimator}
\label{alg:decoupled-st}
\begin{algorithmic}[1]
\STATE \textbf{Input:} logits $\bm{l} \in \mathbb{R}^k$, forward temperature $\tau_f$, backward temperature $\tau_b > 0$
\STATE \textbf{Forward:}
\STATE \quad $\bm{p}^f \leftarrow \text{softmax}(\bm{l} / \tau_f)$ \hfill $\triangleright$ \emph{forward surrogate}
\STATE \quad $\bm{z} \leftarrow \text{discretize}(\bm{p}^f)$ \hfill $\triangleright$ \emph{sample or argmax (if $\tau_f=0)$}
\STATE \quad \textbf{return} $\bm{z}$
\STATE \textbf{Backward:} given $\frac{\partial \mathcal{L}}{\partial \bm{z}}$
\STATE \quad $\bm{p}^b \leftarrow \text{softmax}(\bm{l} / \tau_b)$ \hfill $\triangleright$ \emph{backward surrogate}
\STATE \quad \textbf{return} $\frac{\partial \mathcal{L}}{\partial \bm{l}} \leftarrow \frac{1}{\tau_b} \left( \text{diag}(\bm{p}^b) - \bm{p}^b {\bm{p}^b}^\top \right) \frac{\partial \mathcal{L}}{\partial \bm{z}}$
\end{algorithmic}
\end{algorithm}

\begin{figure*}[htbp]
    \centering
    \begin{subfigure}{0.48\textwidth}
        \centering
        \includegraphics[width=\linewidth]{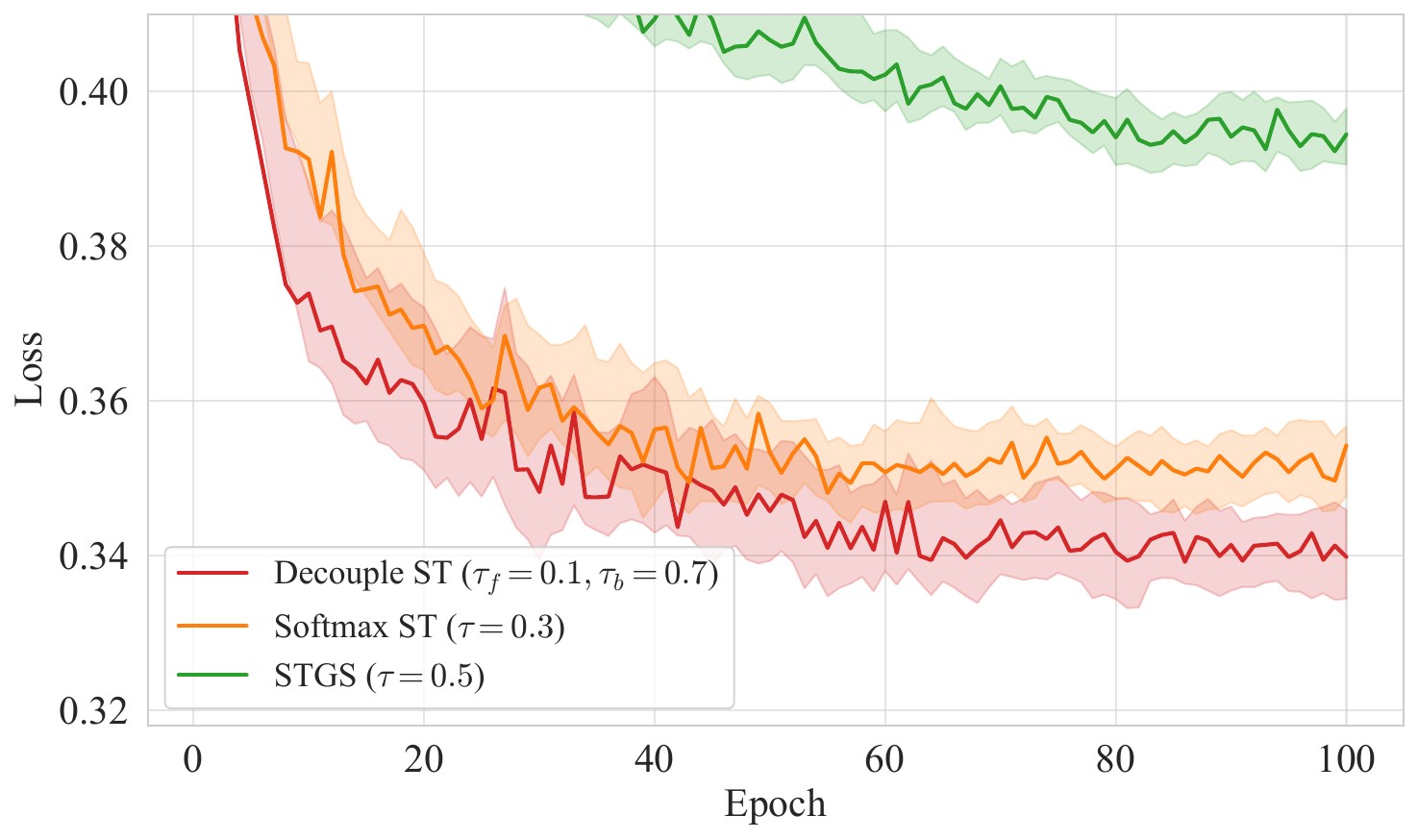}
        \caption{Validation Loss}
    \end{subfigure}
    \hfill
    \begin{subfigure}{0.48\textwidth}
        \centering
        \includegraphics[width=\linewidth]{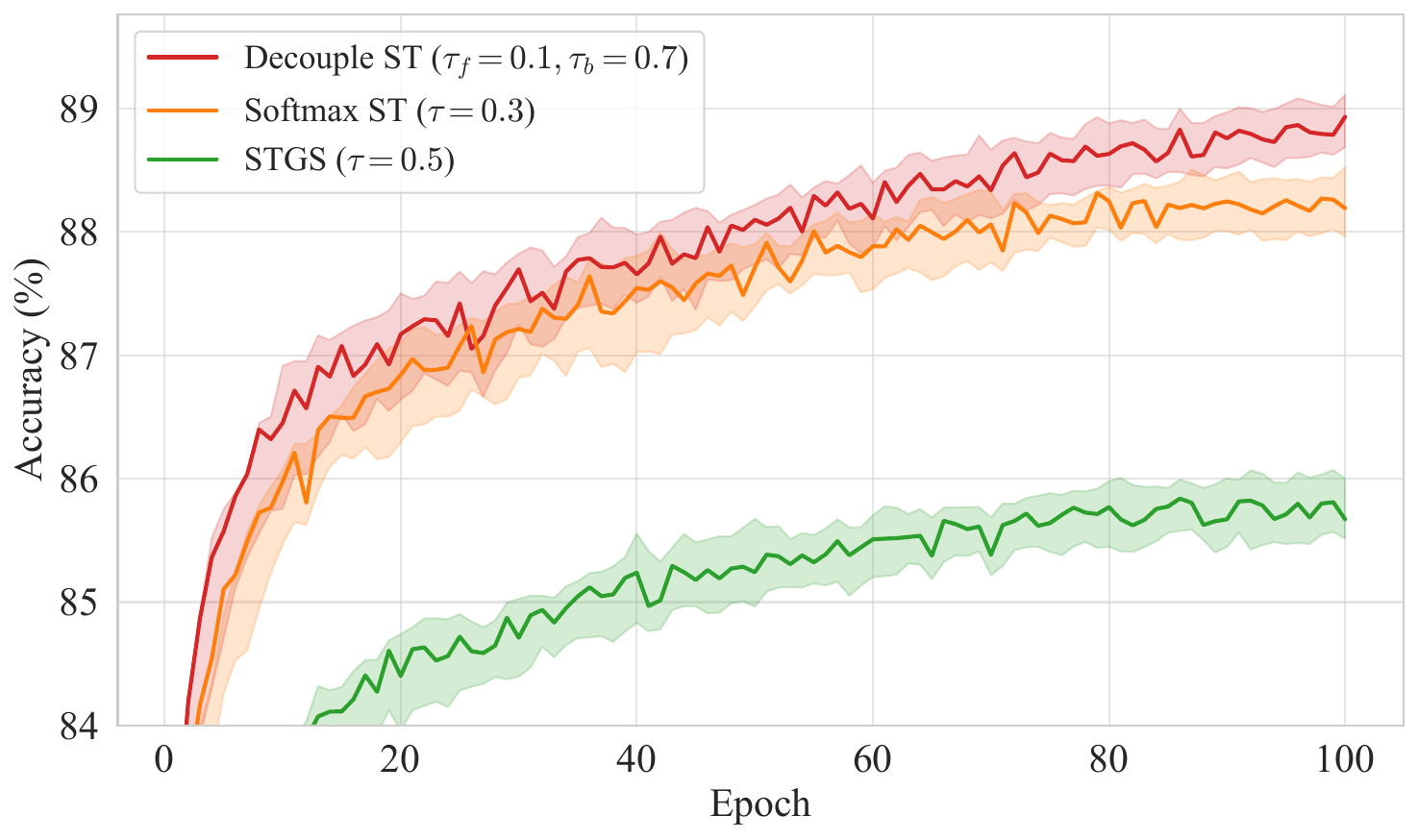}
        \caption{Validation Accuracy}
    \end{subfigure}
    \hfill
    \caption{\textbf{Stochastic Binary Networks on FashionMNIST.} (a, b) Decoupled ST with $\tau_f = 0.1$ and $\tau_b = 0.7$ achieves faster convergence and higher final accuracy than all baselines. The optimal configuration lies off the diagonal ($\tau_f \neq \tau_b$), confirming the benefit of decoupling.}
    \label{fig:sbn_exps}
\end{figure*}

\subsection{Effect of Forward Temperature on Exploration and Variance}

The forward temperature $\tau_f$ controls the entropy of the sampling distribution $\bm{p}^f$. Increasing $\tau_f$ softens the distribution toward uniformity, encouraging exploration and improving latent space utilisation. However, since the gradient estimate (Equation~\ref{eq:decoupled-gradient}) depends on the stochastic sample $\bm{z} \sim \text{Categorical}(\bm{p}^f)$, the variance of the estimator is directly affected by the sampling distribution.

To see this, consider the expected gradient and its variance:
\begin{equation}
    \mathbb{E}_{\bm{z} \sim \bm{p}^f}\left[ \hat{g}(\bm{l}) \right] = J(\tau_b)^\top \mathbb{E}_{\bm{z}}\left[ \frac{\partial \mathcal{L}(\bm{z})}{\partial \bm{z}} \right].
\end{equation}
The variance of the gradient estimate arises from the randomness in $\bm{z}$:
\begin{equation}
    \text{Var}_{\bm{z} \sim \bm{p}^f}\left[ \hat{g}(\bm{l}) \right] = J(\tau_b)^\top \text{Cov}_{\bm{z}}\left[ \frac{\partial \mathcal{L}(\bm{z})}{\partial \bm{z}} \right] J(\tau_b).
\end{equation}
When $\tau_f$ is high, the distribution $\bm{p}^f$ approaches uniformity, meaning each category is sampled with similar probability. This increases the variability in $\bm{z}$ across training steps, which in turn increases the variance of $\partial \mathcal{L}(\bm{z}) / \partial \bm{z}$ and hence the overall gradient variance. Conversely, when $\tau_f$ is low, $\bm{p}^f$ concentrates on one category, reducing sampling variability but limiting exploration.

The forward temperature thus mediates an \emph{exploration-variance trade-off}: high $\tau_f$ improves latent utilisation at the cost of noisier gradients.

\subsection{Effect of Backward Temperature on Gradient Magnitude}

The backward temperature $\tau_b$ controls how the softmax Jacobian reweights gradients across categories. From the Jacobian $J(\tau_b) = \frac{1}{\tau_b}\big( \text{diag}(\bm{p}^b) - \bm{p}^b {\bm{p}^b}^\top \big)$, the overall gradient magnitude scales inversely with $\tau_b$:
\begin{equation}
    \left\| \hat{g}(\bm{l}) \right\| \propto \frac{1}{\tau_b}.
\end{equation}
At low $\tau_b$, the backward softmax $\bm{p}^b$ concentrates sharply on the highest-logit category. This provides strong gradients to the selected category but near-zero gradients to alternatives, risking ``dead categories'' that never receive meaningful updates. At high $\tau_b$, $\bm{p}^b$ approaches a uniform distribution, spreading gradients more evenly across all categories but diluting the learning signal for each individual category.

The backward temperature thus mediates a \emph{magnitude-dispersion trade-off}: low $\tau_b$ provides strong but concentrated gradients; high $\tau_b$ provides weaker but more dispersed gradients.

\subsection{Why Decoupling is Necessary}

Each temperature has its own ``Goldilocks zone'' where performance is optimized, but these zones generally do not coincide. For $\tau_f$, the optimal zone balances exploration against gradient variance: too low causes underutilization of the latent space, while too high destabilizes training. For $\tau_b$, the optimal zone balances gradient magnitude against dispersion: too low causes dead categories, while too high dilutes learning signals.

Crucially, these trade-offs are largely independent. A task that requires high exploration (e.g., autoencoders with large codebooks) benefits from high $\tau_f$, but may still need moderate $\tau_b$ to maintain gradient quality. Conversely, a task where gradient dead-zones are the primary concern (e.g., binary networks) may benefit from high $\tau_b$ to spread gradients, while keeping $\tau_f$ low for stable training.

Any single-temperature method constrains the operating point to the diagonal $\tau_f = \tau_b$, forcing a compromise between these independent concerns. Our experiments (Section~\ref{sec:experiments}) confirm that optimal configurations consistently lie off this diagonal across diverse tasks, demonstrating that the two questions indeed require different answers.

\begin{figure*}[htbp]
    \centering
    \begin{subfigure}{0.48\textwidth}
        \centering
        \includegraphics[width=\linewidth]{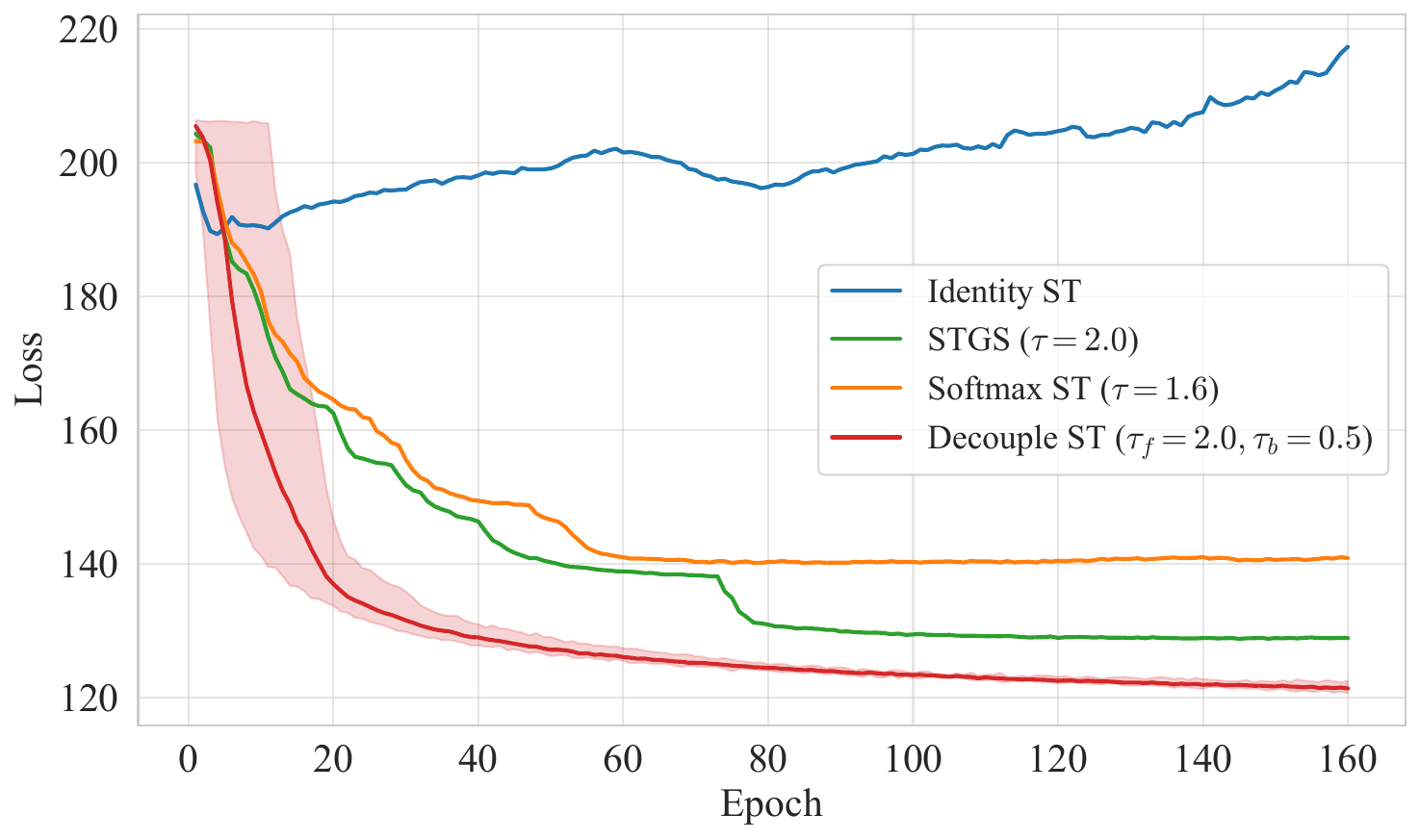}
        \caption{Validation Loss}
    \end{subfigure}
    \hfill
    \begin{subfigure}{0.48\textwidth}
        \centering
        \includegraphics[width=\linewidth]{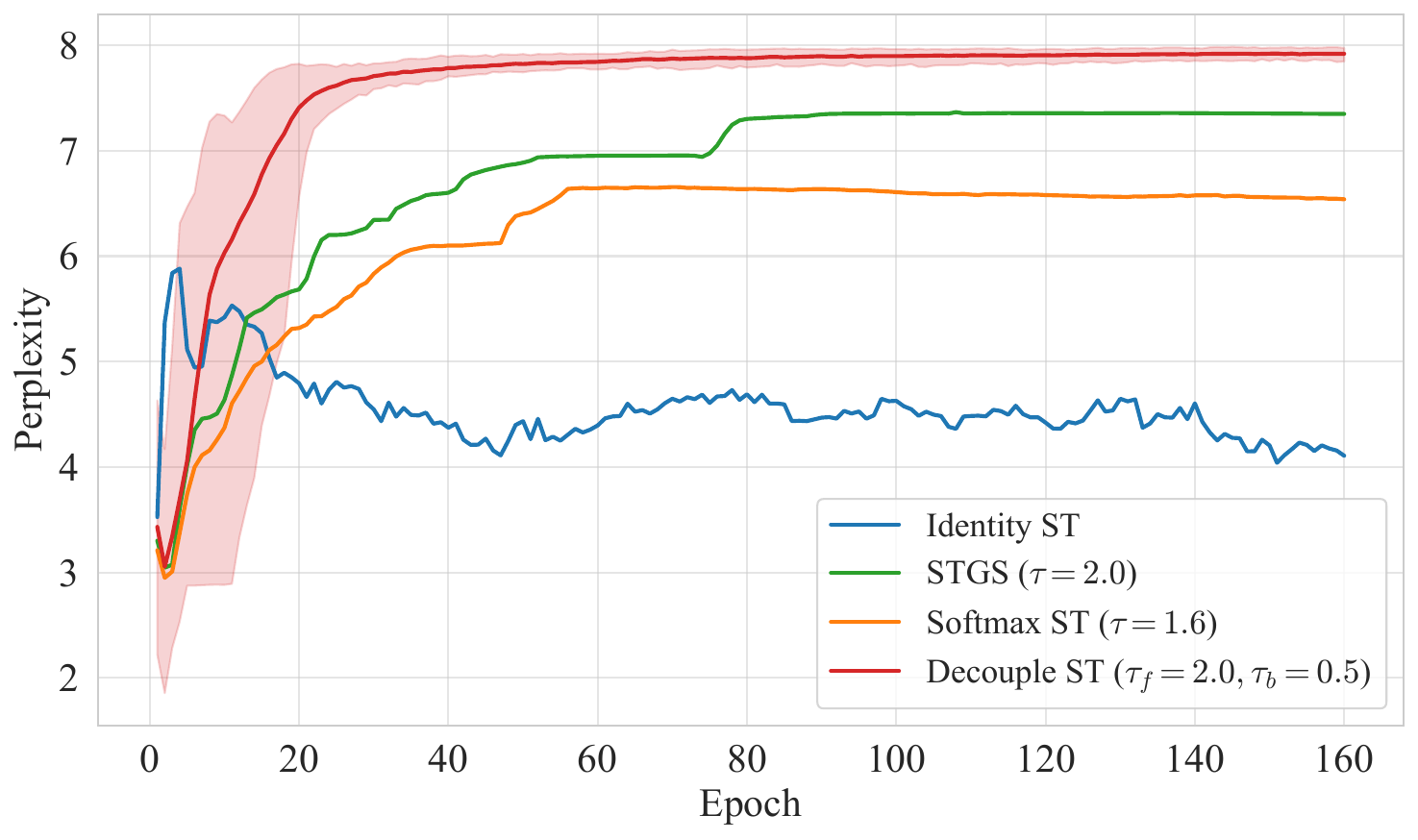}
        \caption{Latent Space Utilization}
    \end{subfigure}
    \hfill
    \caption{\textbf{Categorical Autoencoder on MNIST (4 latents $\times$ 8 classes).} (a) Decoupled ST achieves the lowest validation loss and consistent convergence across all seeds, whereas baselines fail to converge reliably despite extensive temperature and learning rate tuning (error bands omitted due to high variance). (b) Latent space utilization, measured by perplexity (the effective number of codes used), shows that Decoupled ST achieves near-maximum utilization ($\approx 7.9$ out of 8), while baselines severely underutilize the available codes.}
    \label{fig:catae_exps}
\end{figure*}

\section{Experiments and Results}
\label{sec:experiments}

We evaluate Decoupled ST on three tasks spanning different discrete bottleneck types: \textbf{(1)} Stochastic Binary Networks (SBNs) for classification, \textbf{(2)} Categorical Autoencoders for reconstruction, and \textbf{(3)} Differentiable Logic Gate Networks (DLNs) for discrete inference. We compare against three baselines: \textbf{Identity STE}, \textbf{Softmax STE}, and \textbf{Straight-Through Gumbel-Softmax (ST-GS)}. All experiments use the Adam optimiser, and both Softmax STE and ST-GS undergo temperature tuning.
\subsection{Performance Analysis}

\subsubsection{Stochastic Binary Networks}

Stochastic Binary Networks (SBNs) use binary activations sampled from Bernoulli distributions, creating an information bottleneck that can improve generalisation and computational efficiency. The key challenge is gradient dead-zones: binary neurons that receive near-zero gradients become ``stuck'' and stop learning. This task tests whether decoupling can address the magnitude-dispersion trade-off discussed in Section~\ref{sec:methodology}.

We train a fully-connected network with two hidden layers of 200 binary neurons each on FashionMNIST for 100 epochs (batch size 128, learning rate 0.001, averaged over 5 seeds). For Decoupled ST, we perform a full grid search over $\tau_f \in \{0.0, 0.1, \ldots, 2.0\}$ and $\tau_b \in \{0.1, 0.3, \ldots, 2.0\}$, selecting the configuration with the best validation loss. Configurations along the diagonal ($\tau_f = \tau_b$) recover standard Softmax STE.

The optimal configuration is $\tau_f = 0.1$ and $\tau_b = 0.7$, which lies off the diagonal. As shown in Figure~\ref{fig:sbn_exps}(a, b), Decoupled ST outperforms all baselines in both convergence speed and final accuracy. The low forward temperature reduces sampling variance for stable training, while the moderate backward temperature ensures gradients reach inactive neurons. Notably, at training end, only 11\% of neurons are inactive (receiving negligible gradients) for Decoupled ST, compared to 47\% for Softmax STE and 60\% for ST-GS.

While the grid search provides a comprehensive view of the temperature landscape, it requires $O(n^2)$ evaluations and becomes infeasible for larger-scale experiments. For the remaining tasks, we adopt a sequential tuning strategy that requires only $O(n)$ evaluations: first sweep $\tau_f$ with $\tau_b$ fixed, then sweep $\tau_b$ with $\tau_f$ fixed at its best value. Even with this more efficient search, Decoupled ST consistently outperforms baselines.

\subsubsection{Categorical Autoencoder}

Categorical autoencoders compress inputs into discrete latent codes, where the primary challenge is latent space utilisation: models often collapse to using only a small subset of available codes. This task tests whether decoupling can address the exploration-variance trade-off, since high stochasticity is needed to explore codes but can destabilise training.

We train on MNIST with 4 categorical latents, each with 8 classes ($8^4 = 4096$ possible codes), using binary cross-entropy loss for 160 epochs (batch size 200, averaged over 10 seeds). Using the sequential tuning strategy, we first fix $\tau_b = 1.0$ and sweep $\tau_f$, then fix $\tau_f$ at its best value (based on validation loss) and sweep $\tau_b$. All configurations are additionally tuned over learning rates in $\{0.0003, 0.0005, 0.0007, 0.001\}$.

The optimal configuration is $\tau_f = 2.0$ and $\tau_b = 0.5$. As shown in Figure~\ref{fig:catae_exps}(a), Decoupled ST converges consistently across all seeds while \emph{none of the baselines converge reliably}, even after extensive temperature and learning rate tuning. Furthermore, Decoupled ST achieves near-maximum latent utilisation (perplexity $\approx 7.9$ out of 8), while baselines collapse to a small subset of codes (Figure~\ref{fig:catae_exps}b). The optimal configuration lies far from the diagonal: high $\tau_f$ encourages exploration, while lower $\tau_b$ provides strong gradient signals.

\begin{figure*}[htbp]
    \centering
    \begin{subfigure}{0.5\textwidth}
        \centering
        \includegraphics[width=\linewidth]{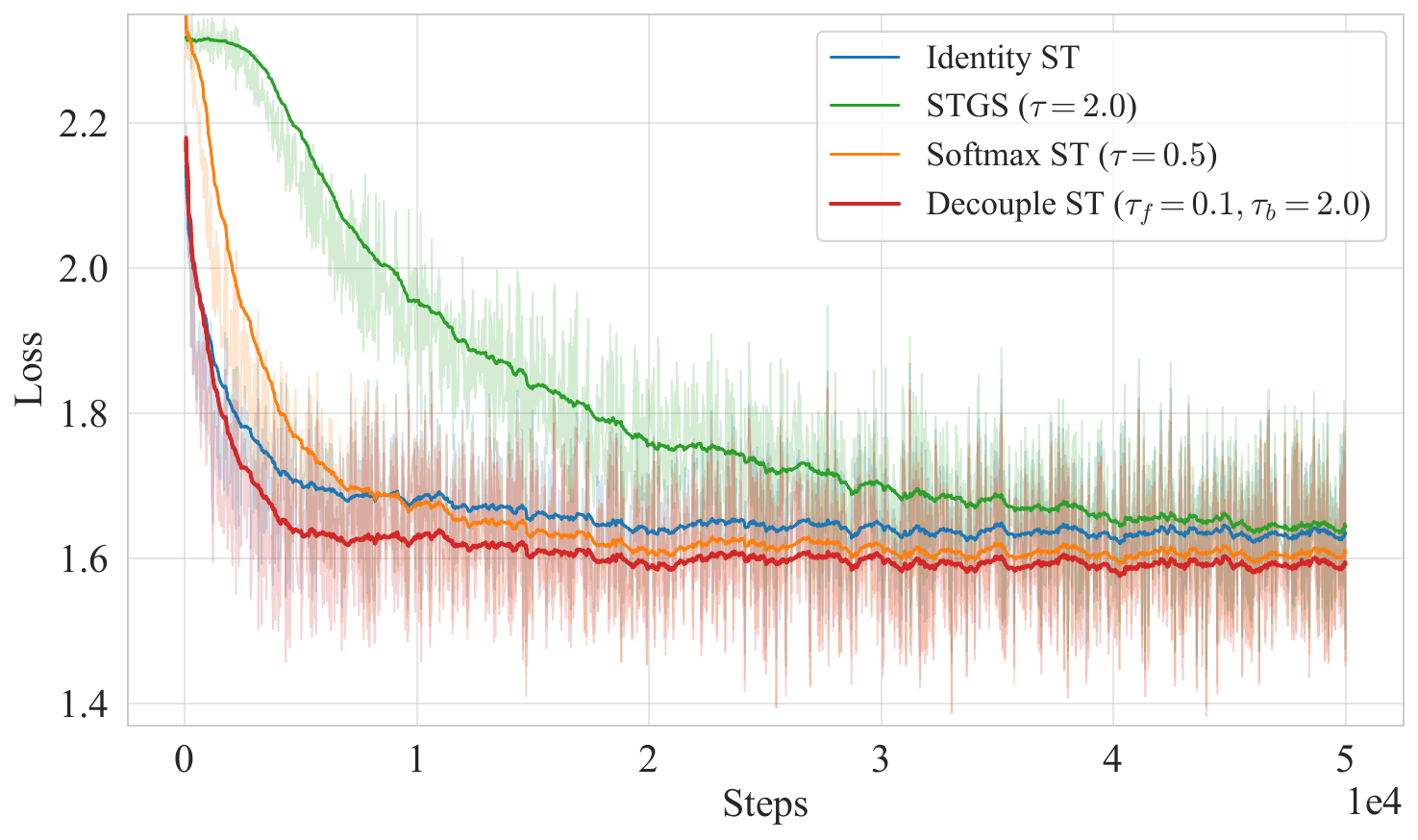}
        \caption{Training Loss}
    \end{subfigure}
    \hfill
    \begin{subfigure}{0.46\textwidth}
        \centering
        \includegraphics[width=\linewidth]{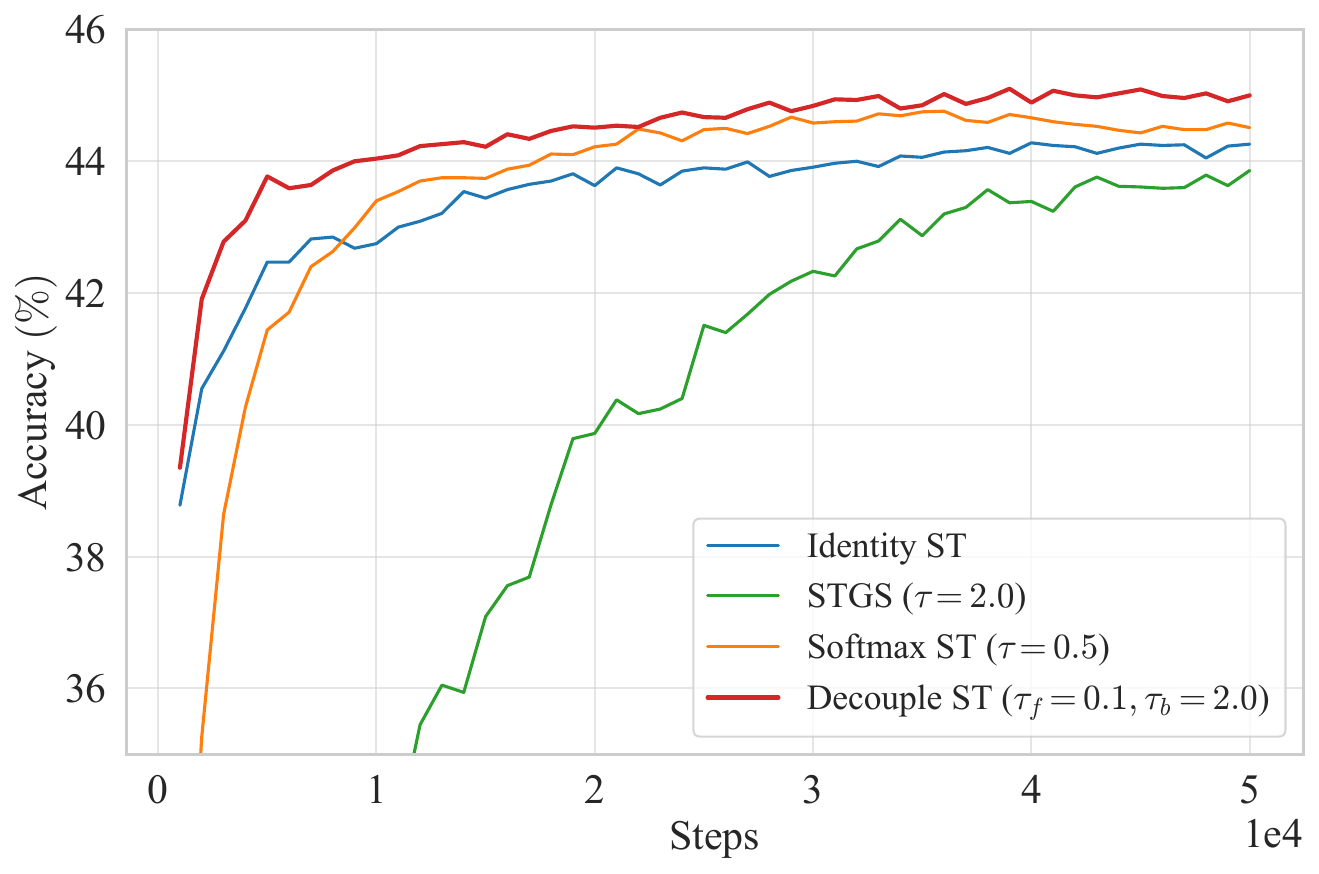}
        \caption{Validation Accuracy}
    \end{subfigure}
    \hfill
    \caption{\textbf{Differentiable Logic Gate Networks on CIFAR10.} (a, b) Decoupled ST with $\tau_f = 0.1$ and $\tau_b = 2.0$ achieves faster convergence and higher final accuracy than all baselines. The low forward temperature encourages near-deterministic gate selection for stable training, while the high backward temperature ensures broad gradient flow across gate options.}
    \label{fig:dln_exps}
\end{figure*}

\subsubsection{Differentiable Logic Gate Networks}

\begin{figure}[t]
  \centering
  \includegraphics[width=\columnwidth]{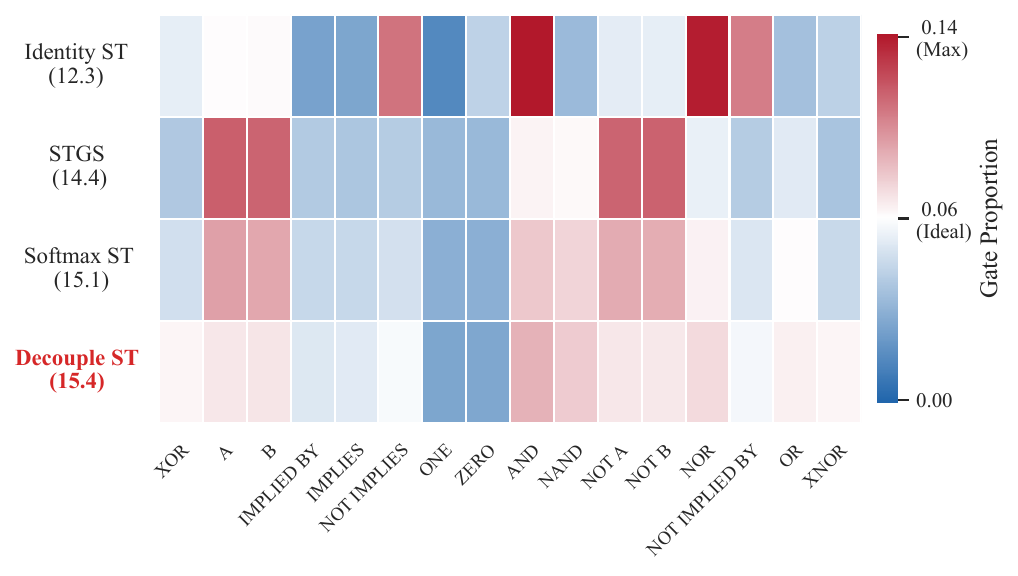}
  \caption{\textbf{Gate distribution in Differentiable Logic Gate Networks.} Distribution across the 16 logic gate types after training, where lighter colors indicate more uniform usage. Decoupled ST achieves the most balanced distribution, measured by perplexity (higher is more uniform, max is 16), indicating effective gradient dispersion across the combinatorial gate space.}
  \label{fig:gate_distro}
\end{figure}

Differentiable Logic Gate Networks (DLNs) \citep{petersen2022deep} compose Boolean logic gates into neural architectures, with each neuron selecting one of 16 possible gate types (AND, OR, XOR, etc.). The combinatorial gate space creates a challenging discrete optimisation problem where dead gates (gates that never receive gradient updates) severely limit expressivity. This task tests whether decoupling can maintain gradient flow across a large discrete action space.

We train a 4-layer DLN with 12,000 neurons per layer on CIFAR10 (binarised inputs) for 200,000 iterations (batch size 100). Using the same two-stage search protocol, the optimal configuration is $\tau_f = 0.1$ and $\tau_b = 2.0$. As shown in Figure~\ref{fig:dln_exps}(a, b), Decoupled ST achieves superior convergence and accuracy. The low forward temperature encourages near-deterministic gate selection for stable inference, while the high backward temperature ensures gradients flow broadly across all gate options. Figure~\ref{fig:gate_distro} shows that Decoupled ST achieves the most balanced gate distribution (highest perplexity), indicating effective exploration of the combinatorially large gate space.

\subsection{Analysing the Effect of Both Temperatures}

Having established the performance benefits of Decoupled ST, we now investigate the distinct roles of $\tau_f$ and $\tau_b$ through empirical analysis.

\begin{figure*}[htbp]
    \centering
    \begin{subfigure}{0.48\textwidth}
        \centering
        \includegraphics[width=\linewidth]{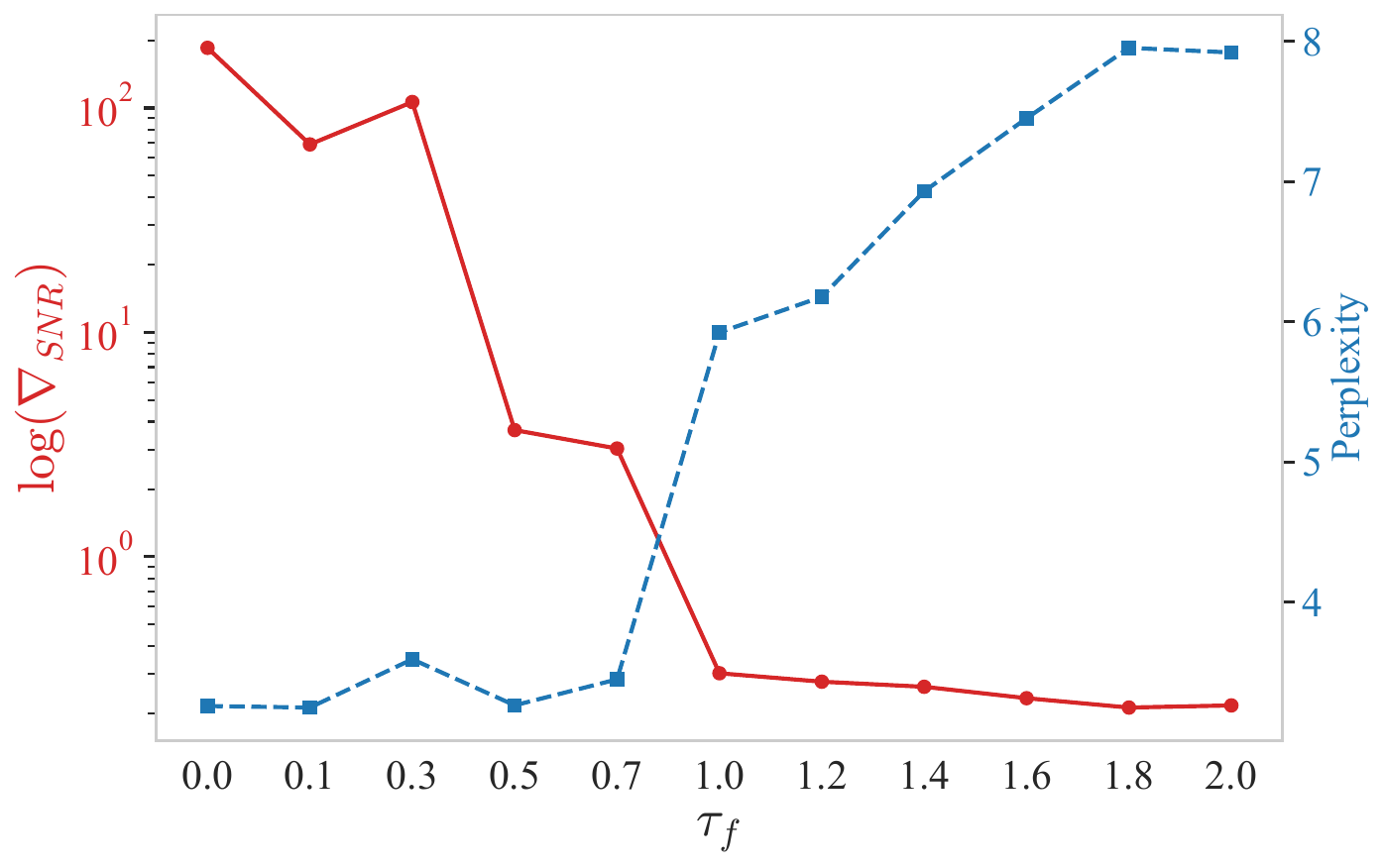}
        \caption{Effect of $\tau_f$}
    \end{subfigure}
    \hfill
    \begin{subfigure}{0.48\textwidth}
        \centering
        \includegraphics[width=\linewidth]{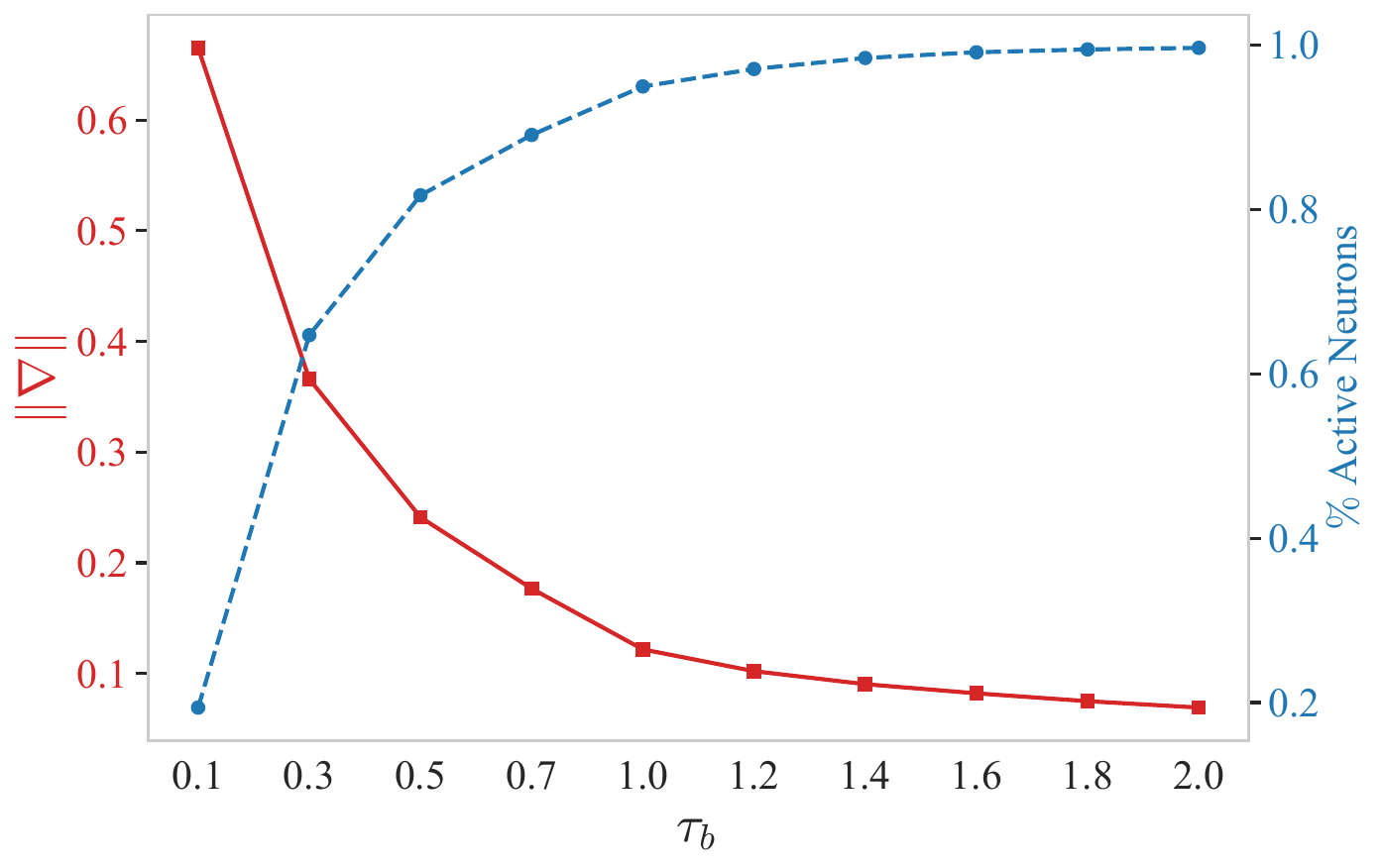}
        \caption{Effect of $\tau_b$}
    \end{subfigure}
    \hfill
    \caption{\textbf{Temperature effects on training dynamics.} (a) $\tau_f$ mediates an exploration-variance trade-off: increasing $\tau_f$ improves latent utilisation (blue) but worsens gradient signal-to-noise ratio (red). (b) $\tau_b$ mediates a magnitude-dispersion trade-off: gradient norm (blue) decreases with increasing $\tau_b$, while the percentage of active neurons receiving updates (red) increases.}
    \label{fig:temp_effects}
\end{figure*}

\subsubsection{Forward Temperature: Exploration vs.\ Gradient Variance}

The forward temperature $\tau_f$ controls sampling stochasticity, mediating a trade-off between \emph{exploration} and \emph{gradient variance}. To quantify gradient noise, we compute the gradient signal-to-noise ratio (SNR): for each training step, we perform 1024 independent forward-backward passes and compute $\text{SNR} = |\mu(\nabla)| / \sigma(\nabla)$, the ratio of mean gradient magnitude to standard deviation.

Figure~\ref{fig:temp_effects}(a) shows the effect of $\tau_f$ on the Categorical Autoencoder task. As $\tau_f$ increases, gradient SNR decreases (more noise), while latent space utilisation increases (more exploration). The optimal $\tau_f$ balances these competing objectives: it must be high enough for adequate exploration but low enough to maintain stable gradient signals.

\subsubsection{Backward Temperature: Gradient Magnitude vs.\ Dispersion}

The backward temperature $\tau_b$ controls gradient fidelity, mediating a trade-off between \emph{gradient magnitude} and \emph{gradient dispersion}. Figure~\ref{fig:temp_effects}(b) shows the effect of $\tau_b$ on gradient statistics in SBNs. As $\tau_b$ increases, gradient norm decreases (weaker signals), while the percentage of active neurons receiving non-negligible updates increases (broader dispersion). The optimal $\tau_b$ provides gradients that are strong enough for effective learning yet distributed enough to prevent dead neurons.

These analyses confirm that forward and backward temperatures control fundamentally different aspects of training, explaining why optimal configurations consistently lie off the diagonal $\tau_f = \tau_b$.

\section{Conclusion}

We have identified a fundamental limitation in existing Straight-Through Estimator methods: they conflate forward-pass stochasticity, which governs exploration and latent space utilisation, with backward-pass gradient fidelity, which determines how learning signals are distributed across categories. These two concerns have distinct and often opposing requirements, yet current approaches either bind them to a single temperature parameter or ignore them entirely.

To address this limitation, we proposed Decoupled Straight-Through (Decoupled ST), a minimal modification that introduces separate temperatures for the forward pass ($\tau_f$) and the backward pass ($\tau_b$). This simple change enables independent control over exploration and gradient dispersion without adding architectural complexity. Across three diverse tasks (Stochastic Binary Networks, Categorical Autoencoders, and Differentiable Logic Gate Networks), Decoupled ST consistently outperformed Identity STE, Softmax STE, and Straight-Through Gumbel-Softmax. Notably, the optimal configurations lie far off the diagonal $\tau_f = \tau_b$, empirically confirming that forward stochasticity and backward gradient shaping require fundamentally different settings. Our analysis further demonstrated that $\tau_f$ mediates a trade-off between exploration and variance, improving latent utilisation at the cost of noisier gradients, while $\tau_b$ mediates a trade-off between magnitude and dispersion, balancing gradient strength against the risk of dead categories.

Several directions remain for future work. First, while we employed grid search and sequential tuning to identify optimal temperature configurations, developing adaptive or learnable temperature schedules could reduce the hyperparameter burden. Second, extending Decoupled ST to larger-scale discrete models, such as vector-quantised autoencoders with large codebooks or discrete diffusion models, may reveal further benefits. Finally, understanding the theoretical relationship between the decoupled temperatures and the bias and variance of the STE gradient estimator could provide principled guidance for temperature selection.

\section*{Impact Statement}
This paper presents work whose goal is to advance the field of Machine
Learning. There are many potential societal consequences of our work, none
which we feel must be specifically highlighted here.

% In the unusual situation where you want a paper to appear in the
% references without citing it in the main text, use \nocite

\bibliography{main}
\bibliographystyle{icml2026}

%%%%%%%%%%%%%%%%%%%%%%%%%%%%%%%%%%%%%%%%%%%%%%%%%%%%%%%%%%%%%%%%%%%%%%%%%%%%%%%
%%%%%%%%%%%%%%%%%%%%%%%%%%%%%%%%%%%%%%%%%%%%%%%%%%%%%%%%%%%%%%%%%%%%%%%%%%%%%%%
% APPENDIX
%%%%%%%%%%%%%%%%%%%%%%%%%%%%%%%%%%%%%%%%%%%%%%%%%%%%%%%%%%%%%%%%%%%%%%%%%%%%%%%
%%%%%%%%%%%%%%%%%%%%%%%%%%%%%%%%%%%%%%%%%%%%%%%%%%%%%%%%%%%%%%%%%%%%%%%%%%%%%%%
\newpage
\appendix
\onecolumn

\begin{table*}[t]
\centering
\small
\setlength{\tabcolsep}{4pt}
\section{Claim-Evidence Table}
\caption{\textbf{Summary of claims and supporting evidence.} Each claim is supported by specific experimental results presented in this paper.}
\begin{tabular}{p{0.04\linewidth} p{0.38\linewidth} p{0.38\linewidth} p{0.14\linewidth}}
\hline
\textbf{ID} & \textbf{Claim} & \textbf{Evidence} & \textbf{Metric} \\
\hline
C1 &
Existing STE methods conflate forward-pass stochasticity with backward-pass gradient fidelity, constraining the operating point to $\tau_f = \tau_b$. Optimal configurations lie off this diagonal. &
Grid search over $(\tau_f, \tau_b)$ yields off-diagonal optima across all three tasks: SBN ($\tau_f{=}0.1$, $\tau_b{=}0.7$), CatAE ($\tau_f{=}2.0$, $\tau_b{=}0.5$), DLN ($\tau_f{=}0.1$, $\tau_b{=}2.0$). &
Validation loss, accuracy \\

C2 &
Decoupled ST consistently improves performance across diverse discrete models. &
Decoupled ST outperforms Identity STE, Softmax STE, and ST-GS on SBN, CatAE, and DLN in both convergence speed and final performance. &
Accuracy, reconstruction loss \\

C3 &
$\tau_f$ and $\tau_b$ control fundamentally different aspects of training: $\tau_f$ mediates an exploration-variance trade-off, while $\tau_b$ mediates a magnitude-dispersion trade-off. &
CatAE: increasing $\tau_f$ improves perplexity but worsens gradient SNR. SBN: increasing $\tau_b$ reduces gradient norm but increases active neuron percentage. &
Perplexity, gradient SNR, gradient norm, \% active neurons \\

\hline
\end{tabular}
\label{tab:claim_evidence}
\end{table*}

%%%%%%%%%%%%%%%%%%%%%%%%%%%%%%%%%%%%%%%%%%%%%%%%%%%%%%%%%%%%%%%%%%%%%%%%%%%%%%%
%%%%%%%%%%%%%%%%%%%%%%%%%%%%%%%%%%%%%%%%%%%%%%%%%%%%%%%%%%%%%%%%%%%%%%%%%%%%%%%

\end{document}